\documentclass[sigconf, nonacm]{acmart}
\AtBeginDocument{%
  \providecommand\BibTeX{{%
    \normalfont B\kern-0.5em{\scshape i\kern-0.25em b}\kern-0.8em\TeX}}}

\setcopyright{acmlicensed}
\copyrightyear{2024}
\acmYear{2024}
\acmDOI{XXXXXXX.XXXXXXX}

\acmConference[End-User Development for Human-Robot Interaction]{Make sure to enter the correct
  conference title from your rights confirmation emai}{March 15,
  2024}{Boulder, CO}
\usepackage{multicol}
\begin{document}

\newcommand{\yuna}[1]{{\color{red}\textbf{Yuna:} #1}}

\title{Understanding Generative AI in Robot Logic Parametrization}

\author{Yuna Hwang}
\orcid{0000-0001-7726-8003}
\email{yunahwang@cs.wisc.edu}
\affiliation{%
  \institution{University of Wisconsin-Madison}
  \streetaddress{1210 W.Dayton Street}
  \city{Madison}
  \state{Wisconsin}
  \country{USA}
  \postcode{53715}
}

\author{Arissa J. Sato}
\orcid{0000-0002-1103-8050}
\email{asato@wisc.edu}
\affiliation{%
  \institution{University of Wisconsin-Madison}
  \streetaddress{1210 W.Dayton Street}
  \city{Madison}
  \state{Wisconsin}
  \country{USA}
  \postcode{53715}
}

\author{Pragathi Praveena}
\orcid{0000-0002-1103-8050}
\email{pragathi@cmu.edu}
\affiliation{%
  \institution{Carnegie Mellon University}
  \streetaddress{1210 W.Dayton Street}
  \city{Pittsburgh}
  \state{Pennsylvania}
  \country{USA}
  \postcode{53715}
}

\author{Nathan Thomas White}
\orcid{0009-0000-9414-9647}
\email{ntwhite@wisc.edu}
\affiliation{%
  \institution{University of Wisconsin-Madison}
  \streetaddress{1210 W.Dayton Street}
  \city{Madison}
  \state{Wisconsin}
  \country{USA}
  \postcode{53715}
}

\author{Bilge Mutlu}
\orcid{0000-0002-9456-1495}
\email{bilge@cs.wisc.edu}
\affiliation{%
  \institution{University of Wisconsin-Madison}
  \streetaddress{1210 W.Dayton Street}
  \city{Madison}
  \state{Wisconsin}
  \country{USA}
  \postcode{53715}
}

\renewcommand{\shortauthors}{Hwang et al.}

\begin{abstract}
Leveraging generative AI (\textit{e.g.,} Large Language Models) for language understanding within robotics opens up possibilities for LLM-driven robot end-user development (EUD). Despite the numerous design opportunities it provides, little is understood about how this technology can be utilized when constructing robot program logic. In this paper, we outline the background in capturing natural language end-user intent and summarize previous use cases of LLMs within EUD. Taking the context of filmmaking as an example, we explore how a cinematography practitioner’s intent to film a certain scene (1) can be articulated using natural language, (2) can be captured by an LLM, and (3) further be parametrized as low-level robot arm movement using an LLM. We explore the capabilities of an LLM interpreting end-user intent and mapping natural language to pre-defined, cross-modal data in the process of iterative program development.
We conclude by suggesting future opportunities for domain exploration beyond cinematography to support language-driven robotic camera navigation. 

\end{abstract}




\begin{teaserfigure}
    \includegraphics[width=\columnwidth]{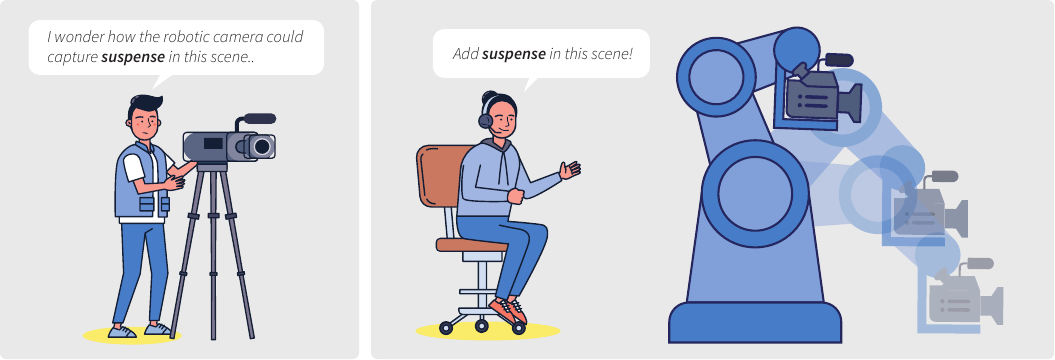}
    \caption{We explore the use of generative AI in robot EUD, specifically robot logic parametrization. One scenario we envision is where generative AI aids a cinematography practitioner in expressing their filming intent and creating robot programs using natural language. \textit{(Left)} While an expert is operating a camera, they imagine how they could use a robotic camera to capture a specific scene. However, this is currently unachievable as there exists a gap between natural language-formatted intent (\textit{i.e.,} how the expert communicates how to capture a scene) and low-level robot behavior (\textit{i.e.,} how the robot receives and interprets the expert's communication). \textit{(Right)} A cinematography practitioner is using \textit{speech} to express their intent instead of operating the robotic camera manually. In this desired interaction flow, the expert can intuitively utilize generative AI (\textit{e.g., } LLMs) in doing so.}
    \label{fig:teaser}
\end{teaserfigure}

\received{20 February 2007}
\received[revised]{12 March 2009}
\received[accepted]{5 June 2009}

\maketitle

\section{Introduction}
Within robot program design, end users have a crucial role in specifying robot behaviors that match unique needs within a domain \cite{barricelli2019end, coronado2019design}. Although end users may have ideas on \textit{what} they want to program within a domain, they may not know \textit{how} to program the robot \cite{chung2020iterative, racca2020interactive}. Here, we motivate this problem by introducing a real-life scenario in which a domain expert uses a robot to perform a specific task. Imagine a filmmaker using a robotic camera to capture a specific scene. The robotic camera could be used as a \textit{cobot}, or a collaborative robot \cite{colgate1996cobots}, to aid an expert in capturing scenes through teleoperation \cite{praveena2023exploring}. Cinematography practitioners may have ideas and intuition on how the scene should be filmed. For example, in a thriller movie, if the expert were to capture \textit{suspense} from a scene, they could have recommendations on what types of shots to use (\textit{e.g., } extreme close-up \cite{heiderich2012cinematography}) to add suspense to the scene. 

The expert would then express their intent of filming the scene using specific techniques. 
However, the question remains: how would they convey their intent if they are not familiar with robot operation methods?

Given the aforementioned example, we identify two problems associated with expert end-users giving instructions to a robot. 
First, end users may not be equipped with the specific semantic rules associated with robot operation. Put differently, end users may not know what a \textit{well-informed intent} \cite{liu2023wants} should look like, making it challenging for users to formulate their intent for effective communication. 
For example, experts may understand how to frame and control the camera movement to match their cinematic vision; however, they may not be as well-trained to verbalize these into commands. 
Experts may face difficulties in going through trial-and-error, framing their intent in different formats of statements such as controlling verbosity level and repeating certain phrases or keywords to emphasize their intent. 
Second, end users’ well-informed intent may not suffice to generate complete, low-level behaviors of a robot \cite{liu2023wants}. 
As with all robotics, factors such as how the robotic parameters (e.g., what signals are sent to the servo motors) should change are essential to the movement of robots. 
However, understanding and adopting the syntactic-level operation rules of a robot is a challenging task for the user \cite{chung2020iterative}, considering that learning how to program a robot requires much training/experience even for experienced programmers \cite{racca2020interactive, ritschel2023training}. 
While experts can familiarize themselves with how to ``talk’’ to the robot, this does not mean they can directly write programs to operate the robot.
Thus, it is necessary to explore how a system can infer how these robotic parameters should change to match the expert's vision.

To address the first problem, we see potential in LLMS as they afford an intuitive way for users to utilize natural language in expressing intent. Although user utterances provided (\textit{i.e., } prompts) for the LLM may not be used verbatim in capturing user intent and implementing programs, this way of communication through natural language allows users to express their programming intent intuitively. 
To address the second problem, we plan to fetch and translate user intent from natural language into parameters that may be used to define a preliminary set of motions for the robotic camera. 
LLMs could be used to refine user intent to generate robot-understandable, well-formed intent by generalizing user utterances and facilitating the modification of low-level behavior parameters by suggesting feasible changes and ranges for low-level behavior parameters using a probabilistic approach.

Our work is heavily inspired by Praveena et al.’s work \cite{praveena2023exploring} that explored the potential use of robotic cameras in the cinematography domain. Within this domain, we (1) explore how LLMs can be utilized as a robot EUD tool in capturing and parametrizing user intent, and (2) how the LLM-driven technique can work towards closing the \textit{abstraction gap} \cite{liu2019review, sarkar2022like, liu2023wants} (\textit{i.e.,} the gap between user-formulated intent and the solution space that is specific to the system).

\begin{figure*}[t]
    \centering
    \includegraphics[width=\textwidth]{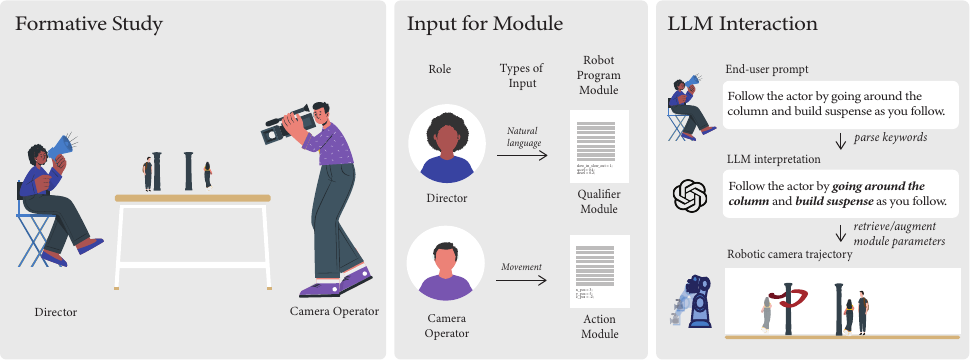}
    \caption{\textit{(Left)} Formative study: We plan to conduct a formative study where a cinematography expert (denoted as the director) uses natural language and movement to express filming intent. The experimenter will act as the camera operator, closely following the expert's instruction and moving the camera. \textit{(Middle)} Input for Module: Different types of input modality will be used to construct separate robot program modules, namely the \textit{action} module and the \textit{intent} module. \textit{(Right)} LLM Interaction: We envision the interaction scenario with the robotic arm mediated by LLM, where it interprets keywords within the end-user prompt and maps those keywords to pre-defined modules and respective parameter values. A robotic camera can initiate a movement given those parameter values.}
    \label{fig:schema}
\end{figure*}
\section{Using Natural Language in Programming Robots}

Natural language allows end users to follow an intuitive communication approach rather than having to speak in a rigid or unfamiliar manner \cite{kollar2010toward}. Researchers have built systems that allow end users to express their intent through natural language \cite{porfirio2023sketching}. Language-based interfaces can also enable the expression of intent within complex tasks \cite{tellex2020robots} and capture underlying semantics of verbal commands with semantic parsing \cite{zettlemoyer2012learning}.
Semantic interpretation has often been utilized to capture program logic and connect aspects of the physical world to the logic statements in which the robot operates (\textit{e.g.,} \cite{matuszek2013learning,  raman2013sorry, artzi2014programming, kshirsagar2019specifying}). 
Liu and Zhang \cite{liu2019review} classified this approach as using \textit{logic models} when understanding user intent and generating execution plans for the robot. 

Some researchers specifically used \textit{probabilistic models} that learn mappings between user commands and action spaces of autonomous robots (\textit{e.g.,} \cite{fasola2013using}, \cite{tellex2011approaching}). 
Researchers also described human-robot interaction (HRI) systems as suites of components that capture high-level cognitive processes. Particularly, the Distributed Integrated Affect Reflection Cognition (DIARC) architecture (\cite{scheutz2013novel}, \cite{scheutz2019overview}) placed the natural language interaction component central to integration with other components for goal managing, action interpretation, and execution. Although literature suggests the potential of probabilistic models in learning semantic associations and emphasizes the role of natural language-based components in coordinating perceptual and action processing, it is still unclear (1) how LLMs can capture user intent and furthermore (2) be used in the robot programming pipeline to support high-level cognitive processes as defined from prior work. \textbf{We aim to discuss LLM's EUD capabilities, as we plan to observe how AI agents interpret end user intent and trigger robot action execution given that LLMs are trained on a vast amount of natural language resources.}


\paragraph{Proposed Scenario} In the filming scenario, LLMs can identify how filmmakers describe scenes using certain keywords. These keywords could be scene-agnostic, \textit{e.g.,} movement-related verbs, spatial prepositions, and names of varying types of shots, or scene-specific, \textit{e.g.,} the object that the filmmaker intends to capture within the scene. Previous work \cite{chiou2023designing} shed light on how ``operational'' expressions (\textit{e.g.,} ``a kid speculating what she sees via a toy camera'') were more effective than ``conceptual'' expressions (\textit{e.g.,} ``ways to speculate'') in generating accurate results within a text-to-image generator (\cite{holz2022midjourney}). However, as both the scene-agnostic and scene-specific keywords defined above fall under the scope of operational expressions, there exists a clear direction to construct a more fine-grained categorization of the expert-provided keywords. Our proposed work includes a formative study that investigates filmmakers' natural language instruction patterns exhibited during scene demonstrations. We expect to acquire clear insight into the different types of keywords employed during natural language interaction, which will then inform us in assessing and utilizing the cognitive capabilities of LLMs.

\section{Generative AI in End-User Development}
Powered by recent advances in deep learning methodologies, generative AI platforms such as ChatGPT\footnote{OpenAI, "ChatGPT 3.5", https://openai.com/chatgpt} and DALL-E\footnote{OpenAI, "DALL-E: Creating Images from Text", https://openai.com/research/dall-e} have showcased success in generating high-fidelity content. Muller et al. envisioned the incorporation of generative AI into creative processes as a collaborative effort between humans and the AI \cite{muller2022genaichi}, and further expanded frameworks on mixed-initiative user interfaces (\textit{e.g.,} \cite{horvitz1999principles, deterding2017mixed, spoto2017library}) to describe human-AI interaction patterns within the generative space \cite{muller2020mixed}. 

Co-creation with AI is an often-visited use case within generative AI-driven EUD. Previous work investigated how end users interact with generative AI when creating different types of media, such as text (\textit{e.g.,} \cite{kim2024understanding, dang2022prompt, yuan2022wordcraft}), images (\textit{e.g.,}  \cite{tseng2024keyframer, liu2022design}), and music (\textit{e.g.,} \cite{louie2020novice, huang2020ai}). 
Generative AI-driven applications can also be found in task planning domains (\textit{e.g., }\cite{shen2024hugginggpt, brohan2023can, driess2023palm, singh2023progprompt}). Systems developed in this domain allow end users to simulate and test their interactions by (1) integrating different computation models without burdening the user and (2) offering realizable solutions in achieving a task goal. Models such as an LLM could operate upon pre-trained semantic knowledge and be further tweaked to provide real-world grounding and feasible \textit{instructions} to meet task goals \cite{brohan2023can}. When prompted to answer high-level user questions such as ``I spilled my drink. Can you help me clean it?’’, \textit{SayCan}~\cite{brohan2023can} can provide reasonable and feasible responses based on real-world context. In their work, the authors utilized LLMs to capture user task goals, break down the tasks into subtasks, and generate narratives that include task solutions \cite{brohan2023can}. Some task-planning problems can be defined more succinctly as a collaboration between AI models \cite{shen2024hugginggpt}. Tasks such as image captioning allow an LLM to be used as a ``task controller,’’ in which it plans the execution of existing HuggingFace AI models\footnote{https://huggingface.co/models}. 

From the perspective of facilitating iterative development of end-user robot programs, further understanding is needed of generative AI’s role within program creation. Detailed examination of domains and task descriptions, other than those in which LLMs are used to seek inspiration (\textit{i.e.,} co-creation) or to generate solutions in task-oriented prompts (\textit{i.e.,} task planning) will offer new interpretations of LLM behavior. 
\textbf{Our proposed work aims to explore the program parametrization capability of LLMs in unique domains where user tasks are less goal-oriented yet sufficiently contextualized to not require machine ideation help. In particular, we start from a set of pre-defined modules that already encapsulate key domain knowledge (to some degree goal-oriented, as solutions can be formulated using only the available resources). On top of this, we aim to understand how LLMs can assist in intent disambiguation and parametrization (an activity that calls for machine support); matching natural language descriptions to modules and their respective parameters. We expect saving parameter information as preambles in subsequent end-user prompts could generate coherent, in-context artifacts throughout the programming process.}


\paragraph{Proposed Scenario}
We propose a filming scenario where a cinematography practitioner and a robot arm are collaborating to film a series of suspenseful scenes. 
In the initial stages, the practitioner describes and demonstrates their vision of how the scene should be shot: ``Let's create a \textit{suspenseful} scene like this,'' and slowly zooms into an object. 
The LLM-powered robotic arm observes the practitioner's speech and filming technique. 
During this interaction, LLMs can be used in robot logic parametrization.
LLMs could have access to a set of pre-defined \textit{modules} that take responsibility for intent interpretation (the \textit{qualifier} module) and robot action execution (the \textit{action} module) and map end-user keywords to the parameters defined within each module. Parameter values fetched from the modules can subsequently be fed into the robot program (\textit{e.g.,} custom-made ROS nodes) and initiate robot arm movement for filming.
A domain-specific system that considers the synchronization patterns emerging from speech and motion holds the potential for constructing a prototype for the robot program. Prior work by Shen et al. \cite{shen2024hugginggpt} portrays a similar pipeline to what we envision in our proposed work. Here, however, instead of configuring independent HuggingFace AI models, we plan to utilize domain-specific modules for the LLM to search for relevant information within parameter fields. This process is represented in Figure \ref{fig:schema}.

\section{Domain Exploration}
Beyond applications within cinematography, researchers are exploring how speech and robotic arms can be used to guide the filming of detailed tasks, such as shooting tutorials~\cite{li2023stargazer}.
We envision opportunities for generative AI and filming to support other experts in different skill-based domains where visual attention is required for dynamic objects, such as during music lessons or physical therapy sessions.
For example, we can imagine a music learning scenario where a student and music instructor are having an online music lesson. 
The instructor could use natural language (\textit{e.g.,} ``follow my bow arm'') to guide the robotic camera to gain better visuals of the intricate motions of the arm, enhancing the learning experience of the student.
Further exploration into other domains can lead to customized robotic collaborations driven by an expert's speech, providing deeper insights into how generative AI can facilitate domain experts in EUD of robotic cameras.





\section{Conclusion}
This paper summarized natural language-based generative AI applications within EUD and robot programming. We envision LLMs being utilized for robot logic parametrization, specifically for capturing high-level user intent and generating robot programs that have learned, shared semantic representation derived from cross-modal data. In the following research, we plan to collaborate with domain experts in the field of cinematography, observe how they interact with LLMs, and build a prototype system that could be further refined for seamless, and cohesive interaction.

\section*{Acknowledgment}
We thank Amy Koike for her help in creating and refining figures presented in this paper.
The teaser figure is modified from images by jcomp, freepik, storyset, and pikisuperstar on Freepik.

\bibliographystyle{ACM-Reference-Format}
\bibliography{ref}
\end{document}